\documentclass{support/IEEEtran}

\usepackage{mathtools}
\usepackage{graphicx}
\usepackage{epstopdf}
\usepackage{lipsum,adjustbox}
\usepackage{cite}
\usepackage{float}
\usepackage{textcomp}
\usepackage{tikz}
\usepackage[section]{placeins}
\usepackage{subcaption}
\usetikzlibrary{shapes}
\usepackage{comment}

\graphicspath{{figures/}}
\definecolor{yellow}{RGB}{255,255,0}
\definecolor{red}{RGB}{255,91,51}
\definecolor{blue}{RGB}{8,191,223}
\definecolor{green}{RGB}{54,203,30}
\definecolor{grey}{RGB}{170,170,170}
\definecolor{black}{RGB}{0,0,0}
\definecolor{orange}{RGB}{255,140,0}
\definecolor{yellow}{RGB}{225,249,27}
\tikzset{VertexStyle/.style = {shape = rectangle,fill = gray}}

\DeclareMathOperator*{\argmin}{argmin}
\DeclareMathOperator*{\argmax}{argmax}

\begin{document}

\title{ Evaluation of Kinematic Precise Point Positioning Convergence with an Incremental Graph Optimizer \\ }
\author{Ryan M. Watson and Jason N. Gross , { \it West Virginia University} }
\maketitle

\begin{abstract}

 Estimation techniques to precisely localize a kinematic platform with GNSS observables can be broadly partitioned into two categories: differential, or undifferenced. The differential techniques (e.g., real-time kinematic (RTK)) have several attractive properties, such as correlated error mitigation and fast convergence; however, to support a differential processing scheme, an infrastructure of reference stations within a proximity of the platform must be in place to construct observation corrections. This infrastructure requirement makes differential processing techniques infeasible in many locations. To mitigate the need for additional receivers within proximity of the platform, the precise point positioning (PPP) method utilizes accurate orbit and clock models to localize the platform. The autonomy of PPP from local reference stations make it an attractive processing scheme for several applications; however, a current disadvantage of PPP is the slow positioning convergence when compared to differential techniques. In this paper, we evaluate the convergence properties of PPP with an incremental graph optimization scheme (Incremental Smoothing and Mapping (iSAM2)), which allows for real-time filtering and smoothing.  The characterization is first conducted through a Monte Carlo analysis within a simulation environment, which allows for the variations of parameters, such as atmospheric conditions, satellite geometry, and intensity of multipath. Then, an example collected data set is utilized to validate the trends presented in the simulation study.

\end{abstract}

\section{Introduction}

The ability to precisely localize a platform is of paramount importance to a myriad of fields (e.g., augmented reality \cite{augmentedReality}, autonomous navigation \cite{autonomousCar}, and natural hazard monitoring \cite{rosen2006uavsar}). To facilitate the precise localization of the platform, several navigation aids can be utilized (e.g., vision \cite{visionSLAM}, lidar \cite{lidarNav}, inertial \cite{jekeli2001inertial}). One navigation aid that is commonly utilized for terrestrial applications is a global navigation satellite system (GNSS) receiver. The signals propagated by a GNSS satellite provides the algorithm with information that allows for accurate, global localization of the platform.

One commonly used methodology for processing GNSS signals is the Precise Point Positioning (PPP) approach \cite{zumberge1997precise}. The PPP algorithm utilizes the dual-frequency undifferenced GNSS observables, which allows the technique to operate without the need of external reference stations. The undifferenced observations are use along with precise GNSS orbit and clock bias products to mitigate the errors removed through observation differencing \cite{misra2006global}. The orbit and clock products that enable the PPP method to achieve decimeter level positioning can be broadcast to an end-user in real-time (e.g., L-band \cite{dixon2006starfire} and Iridium modem link \cite{muellerschoen2003aviation}).

Real-time kinematic-PPP (kPPP) provides similar positioning performance when compared to traditional differential GPS (DGPS) (i.e., Real Time Kinematic) for dynamic platforms \cite{grossUAV}. The comparable positioning accuracy without the need for a nearby static GPS reference station makes it an attractive processing formulation. However, it has been noted in several studies that the PPP formulation has a longer convergence period than comparable differential techniques \cite{pppConverge1,pppConverge2}.

In an attempt to decrease the initial convergence period, there has been a plethora of research into augmenting the PPP approach with additional information sources. One of the most commonly utilized augmentation sources for traditional single constellation PPP is additional GNSS observables. One example of this type of augmentation is the incorporation of multiple constellation observations \cite{cai2015precise, multiPPP}. Another example of this type of augmentation is the PPP-RTK formulation \cite{pppRTK} which provides faster convergence by enabling integer ambiguity resolution \cite{bertigerAmb}. Another well studied form of PPP augmentation is the tightly coupled PPP inertial navigation (INS) formulation \cite{groves2013gpsins}, which has also been shown to decrease the initial convergence period of PPP \cite{watson2017Flight}. However, all PPP augmented methods require additional infrastructure (e.g., a network of reference stations, or additional sensors on-board), which can be prohibitive for many applications.

Another method to decrease the convergence period of PPP is to utilize a  novel optimization framework. This has the potential to provide a benefit over the previously discussed PPP methods because all the previously provided methods utilizing the same underlying optimization framework (i.e., a variant of the Kalman filter \cite{kalman, julier1997new, bierman2006factorization}). Where this framework estimates the desired states by marginalizing all prior information and propagating with dynamic models to the next time step. For PPP, where a subset of the desired states are not observable over a single epoch (i.e., the carrier-phase ambiguity states), this may not be the best framework.

In this paper we evaluate the convergence properties of PPP utilizing an incremental graph optimization framework that allows for real-time smoothing. This work relies upon advances made within the robotics community on efficient, real-time smoothing. Where research into smoothing, within the robotics community, has been dominated by graph based methodologies since the seminal paper on the subject was published in 1997 \cite{graphSmooth}. When \cite{graphSmooth} was published, graph-based smoothing was not widely utilized due to computation complexity of solving the initial formulation. However, quickly thereafter, methods were proposed to greatly reduce complexity through the utilization of factor graphs \cite{factorGraphOG}. The $\sqrt{SAM}$ formulation as presented in \cite{sam} was particularly influential as it provided connections between the factor graph formulation and sparse linear algebra. The idea of batch factor graph optimization was later extended to an incremental inference framework in \cite{isam, isam2}. The work presented in \cite{isam2} provides a frame-work to conduct real-time \cite{realTime}, non-linear graph based filter and smoothing.

The rest of this paper is organized in the following manner. First, the technical approach will be discussed. The technical discussion will provide an overview of factor graph optimization for GNSS optimization, and the ability to incrementally updating using the Bayes tree data structure. Next the discussion will shift to the evaluation of the algorithm with both simulated and collected datasets. Finally, some concluding remarks and future work will be discussed.

\section{Technical Approach}

\subsection{Factor Graphs}

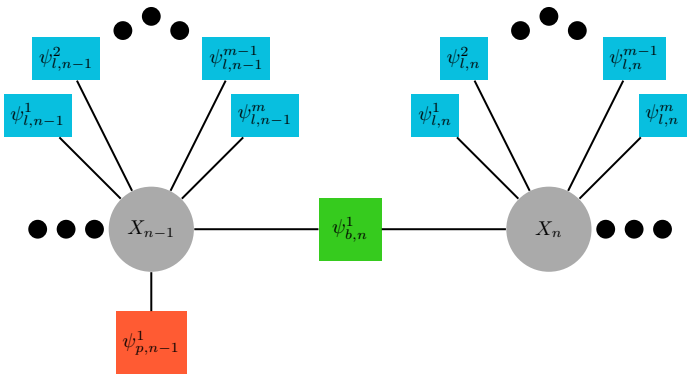
\begin{figure}[htb!]
 \begin{center}
  \begin{adjustbox}{width=0.5\textwidth}
 \begin{tikzpicture}
  \node[shape=circle,fill=black,minimum size=0.1cm] (x) at(2,0) {};
  \node[shape=circle,fill=black,minimum size=0.1cm] (y) at(2.5,0) {};
  \node[shape=circle,fill=black,minimum size=0.1cm] (z) at(3,0) {};

  \node[shape=circle,fill=grey, minimum size=1.5cm] (B) at (4,0) {$X_{n-1}$};
  \node[shape=circle,fill=grey, minimum size=1.5cm] (C) at (11,0) {$X_{n}$};

  \node[shape=rectangle, fill=red, minimum size=1.1cm] (Z) at (4,-2) {$\psi_{p,n-1}^{1}$};

  \node[shape=circle,fill=black,minimum size=0.1cm] (x) at(12,0) {};
  \node[shape=circle,fill=black,minimum size=0.1cm] (y) at(12.5,0) {};
  \node[shape=circle,fill=black,minimum size=0.1cm] (z) at(13,0) {};

  \node[shape=rectangle,fill=blue, minimum size=0.75cm] (H) at (2,2) {$\psi_{l,n-1}^{1}$};
  \node[shape=rectangle,fill=blue, minimum size=0.75cm] (I) at (2.5,3) {$\psi_{l,n-1}^{2}$};
  \node[shape=circle,fill=black,minimum size=0.1cm] (a) at(3.5,3.5) {};
  \node[shape=circle,fill=black,minimum size=0.1cm] (b) at(4,3.75) {};
  \node[shape=circle,fill=black,minimum size=0.1cm] (c) at(4.5,3.5) {};
  \node[shape=rectangle,fill=blue, minimum size=0.75cm] (J) at (5.5,3) {$\psi_{l,n-1}^{m-1}$} ;
  \node[shape=rectangle,fill=blue, minimum size=0.75cm] (K) at (6,2) {$\psi_{l,n-1}^{m}$};

  \node[shape=rectangle,fill=green, minimum size=1.1cm] (Q) at (7.5,0) {$\psi_{b,n}^{1}$};

  \node[shape=rectangle,fill=blue, minimum size=0.75cm] (L) at (9,2) {$\psi_{l,n}^{1}$};
  \node[shape=rectangle,fill=blue, minimum size=0.75cm] (M) at (9.5,3) {$\psi_{l,n}^{2}$};
  \node[shape=circle,fill=black,minimum size=0.1cm] (a) at(10.5,3.5) {};
  \node[shape=circle,fill=black,minimum size=0.1cm] (b) at(11,3.75) {};
  \node[shape=circle,fill=black,minimum size=0.1cm] (c) at(11.5,3.5) {};
  \node[shape=rectangle,fill=blue, minimum size=0.75cm] (N) at (12.5,3) {$\psi_{l,n}^{m-1}$};
  \node[shape=rectangle,fill=blue, minimum size=0.75cm] (O) at (13,2) {$\psi_{l,n}^{m}$};

  \path [-,line width=1pt] (B) edge node {} (Z);
  \path [-,line width=1pt] (B) edge node {} (H);
  \path [-,line width=1pt] (B) edge node {} (I);
  \path [-,line width=1pt] (B) edge node {} (J);
  \path [-,line width=1pt] (B) edge node {} (K);

  \path [-,line width=1pt] (B) edge node {} (Q);
  \path [-,line width=1pt] (Q) edge node {} (C);

  \path [-,line width=1pt] (C) edge node {} (L);
  \path [-,line width=1pt] (C) edge node {} (M);
  \path [-,line width=1pt] (C) edge node {} (N);
  \path [-,line width=1pt] (C) edge node {} (O);

 \end{tikzpicture}
\end{adjustbox}
 \end{center}
 \caption{Example factor graph}
 \label{fig:generic_graph}
\end{figure}

The ability to conduct accurate and efficient inference is at the center of all navigation algorithms. One way to represent the inference problem is with a probabilistic graphical model ~\cite{pearl}, which can take several forms. One convenient graphical model for conducting state estimation is the factor graph ~\cite{factorGraphOG}. At a fundamental level, the factor graph provides a convenient framework for factorizing a function of several variables into smaller subsets. More explicitly, this model provides a useful framework for factorizing the posterior distribution, which allows for efficient calculation of the state vector that maximizes the $a \ posteriori$ distribution. The factorization is represents as a bipartite graph, $\mathcal{G}= (\mathcal{F},\mathcal{X},\mathcal{E})$, where there are two types of vertices: the states to be estimated, $\mathcal{X}$, and the probabilistic constraints applied to those states, $\mathcal{F}$. An edge $\mathcal{E}$ only exists between a state vertex and a factor vertex if that factor is a constraint applied on that time-step. An example factor graph is depicted in Fig.\ref{fig:generic_graph}, where $X_n$ represents the states to be estimate at time-step $n$, $\psi_{p,n-1}$ represents prior information about the estimated states at time-step $n-1$, $\psi_{b,n}$ represents the motion model of the platform from time-step $n-1$ to $n$, and  $\psi_l$ represents the constraint applied to the state by a measurement (e.g., a GNSS pseudorange observable).

As previously mentioned, the factor graph provides a factorization over the posterior distribution, $p(X|Z)$. Thus, we can easily calculate the state vector that maximizes the posterior (MAP) by finding the state vector that maximizes the product of factors, as depicted in Eq. \ref{fg}.

\begin{equation}
 \hat{X} = \argmax_x \ \lbrace \prod_{i=1}^{I} \psi_{p,i}  \prod_{j=1}^{J} \psi_{b,j}  \prod_{k=1}^{K} \psi_{l,k} \rbrace
 \label{fg}
\end{equation}

\noindent For a through discussion on factor graph based state estimation the reader is refered to \cite{dellaert2017factor}.

The optimization problem presented in Eq. \ref{fg} can be reduced to non-linear least-squares formulation if Gaussian noise is assumed, as provided in Eq. \ref{fgGaussian}.

\begin{equation}
 \begin{aligned}
  \hat{X} = \argmin_x  \ \bigg[ & \ \sum_{i=1}^{I} \lvert \lvert x_o - x_i \rvert \rvert^{2}_{\Sigma} \\ & + \sum_{j=1}^{J} \lvert \lvert x_j - f(x_{j-1}) \rvert \rvert^{2}_{\Lambda} \\ & + \sum_{k=1}^{K} \lvert \lvert z_k - h_k(x_{k}) \rvert \rvert^{2}_{\Xi} \bigg]
  \label{fgGaussian}
 \end{aligned}
\end{equation}

Now that a general discussion of the factor graph framework has been provided, we can proceed by constructing GNSS specific factors. For this work, we will detail the construction of two factors: the GNSS observation factor, and the carrier-phase bias factor.

\subsection{Constructing the GNSS Observation Factor}

To allow autonomy of the PPP approach from local reference stations, the undifferenced dual-frequency GNSS observables are utilized. Due to the undifferenced nature of the observations, the PPP processing technique must incorporate GNSS error mitigation models --- these models provide corrects for the corrupting sources that would be mitigated through observation differencing --- to provide an accurate positioning solution. The sources that corrupt a GNSS observation can be segregated into three partitions: the error contributed by the propagation medium, the error contributed by the control segment, and the error contributed by the user.

To begin constructing our measurement model, the method implemented to mitigate the propagation medium errors are discussed. The error attributed to the propagation medium is composed of delay due to the ionosphere and the delay due to the troposphere. To mitigate the ionospheric delay, we leverage the dispersive nature of the medium, and a linear combination of the GPS $L_{1}$ and $L_{2}$ frequencies (1575.42 MHz and 1227.60 MHz, respectively) is formed to produce ionospheric-free (IF) pseudorange and carrier phase measurements ~\cite{misra2006global}. The IF combination of an observable, $O^{j}$, can be seen in Eq. \ref{ionFree}.

\begin{equation}
 O^{j}_{IF}  = O^{j}_{L1}\left[ \frac{f^{2}_{1}}{f^{2}_{1} - f^{2}_{2} } \right] - O^{j}_{L2}\left[ \frac{f^{2}_{2}}{f^{2}_{1} - f^{2}_{2} } \right]
 \label{ionFree}
\end{equation}

To mitigate the error attributed to the troposphere, both the wet and the dry component of the troposphere must be modeled, as shown in Eq. \ref{trop}. For this study, the Hopfield model ~\cite{kaplan2005understanding} is used to model the dry component of the troposphere. To compensate for the wet delay --- the wet component only accounts for approximately 10\% of the total troposphere error --- and the residual error of the dry delay model, a stochastic random variable is added to the state vector.

\begin{equation}
 T(el) = T_{z,d} \mathcal{M}_d(el) + T_{z,w} \mathcal{M}_w(el)
 \label{trop}
\end{equation}

To mitigate the error attributed to the control segment, the PPP approach utilizes orbit and clock corrections. These global corrections are generated through a network of reference stations.

Finally, a discussion on the user error segment is provided. The user error segment is composed of two sources: multipath error, and receiver thermal noise error. For this study, no methods were implemented to explicitly model the user segment error; however, as noted in ~\cite{userError}, the magnitude or the user error is proportional to the elevation angle between the platform and the satellite so, within this evaluation, the uncertainty in the observation is scaled by the elevation angle. It should be noted that PPP observational models for moving platforms typically include corrections for relativistic effects (i.e. from the GPS broadcast correction), receiver and satellite antenna phase center variation, and carrier-phase wind-up; however, these effects were neglected within this simulation study. Additionally, dynamic platform generally couple inertial information with the GNSS observables to mitigate uncertainty in the platforms dynamic model ~\cite{watson2017Flight}.

Utilizing the provided error mitigation techniques, the PPP observation model can be constructed. The pseudorange and carrier-phase measurements are modeled as shown in Eq. \ref{prErr} and Eq. \ref{cpErr}, respectively: where, $R_j=||x^s-x^u||$ is the geometric range between the platform and the $j^{th}$ satellite, $\delta t_{u}$ is the receiver's clock bias, $\delta t_{s}$ is the satellite's clock bias, $T^{}_{z}$ is the tropospheric delay in the zenith direction, $\mathcal{M}_d(el^{j})$ is a user to satellite elevation angle dependent mapping function, $\delta_{Rel.}$ is the correction attributed to relativistic effect \cite{pascual2007introducing}, $\delta_{P.C.}$ is the correction attributed to the offset between the satellite's center of mass and the phase center of the antenna \cite{heroux2001gps}, $\delta_{D.C.B}$ is the differential code bias correction \cite{kaplan2005understanding}, $\delta_{W.U.}$ is the correction attributed to the windup effect on the phase observables \cite{wu1992effects}, $\lambda_{IF}$ is the wavelength corresponding to the IF combination, and $N_{IF}$ is phase ambiguity. In Eqs. \ref{prErr} and \ref{cpErr} the remaining unmodelled error sources are indicated with $\epsilon$. To implement the provided observation model in software, the open-source library GPSTk \cite{harris2007gpstk} is utilized.

\begin{equation}
 \begin{split}
  \rho^{j}_{IF}  = & R_j + c(\delta t^{}_{u} - \delta t_{s})+ T_{z,d}  \mathcal{M}_d(el^{j}) \\
  & + \delta_{Rel.} + \delta_{P.C.} + \delta_{D.C.B} + \epsilon^{j}_\rho
  \label{prErr}
 \end{split}
\end{equation}

\begin{equation}
 \begin{split}
  \boldsymbol{\phi}^{j}_{IF} = & R_j + c(\delta t^{}_{u} - \delta t_{s})+ T_{z,d} \mathcal{M}_{d}(el^{j}) \\
  & + \delta_{Rel.} + \delta_{P.C.} + \delta_{W.U.} + \lambda^{}_{IF} N^{j}_{IF} + \epsilon^{j}_\phi
  \label{cpErr}
 \end{split}
\end{equation}

Using the PPP observation model, we can construct a GNSS constraint for the factor graph \cite{sunderhauf2012multipath}. To begin, we note that the GNSS observations are providing a set of likelihood constraint, $\mathcal{L}(O|X)$, on the optimization process. If the assumption is made that the state and measurement noise models are Gaussian, then this constraint can be incorporated into factor graph through the mahalanobis distance, as provided in Eq. \ref{gnssFactor}, where $z$ is the observed measurement, $\hat{z}$ is the estimated measurement  --- calculated using Eq's \ref{prErr} and \ref{cpErr} --- and $\Sigma$ is the uncertainty in the observation.

\begin{equation}
 \psi_l = (z-\hat{z})^{T} \Sigma^{-1} (z - \hat{z})
 \label{gnssFactor}
\end{equation}

\subsection{Incorporating the Carrier-Phase Ambiguity States}

\begin{figure}[htb!]
 \begin{center}
  \includegraphics[width=0.5\textwidth]{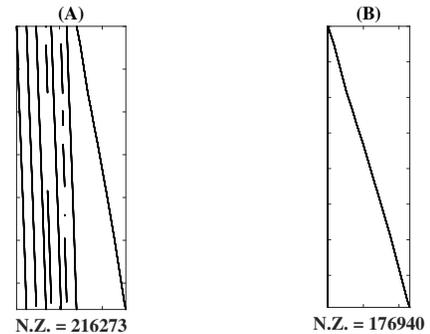}
 \end{center}
 \caption{Sparse measurement Jacobian for the PPP processing strategy. Figure (A) shows the measurement Jacobian when a new carrier-phase ambiguity state is added for each epoch. Figure (B) shows the measurement Jacobian when a new carrier-phase ambiguity is added only when a new satellite is tracked or if a carrier-phase cycle-slip occurs.}
 \label{fig:jacobian}
\end{figure}

There are several way in which the carrier-phase ambiguity states can be incorporated into the factor graph. One such way is to incorporate a new carrier-phase ambiguity state for each epoch. Consecutive carrier-phase ambiguity states can be constrained by a process noise update. The measurement Jacobian associated with this graph construction is represented in Fig. \ref{fig:jacobian}.A. From Fig \ref{fig:jacobian}.A, we can see a measurement Jacobian that is more densely populated than desired.

To construct less densly populated measurement Jacobian (i.e., a more efficient optimization scheme), we can leverage the knowledge that the true carrier-phase ambiguity value for a given satellite within a continually tracked phase-arc is a constant value. Due to this property of the true ambiguity value, the carrier-phase ambiguity factor can be represented as a random constant variable. Where, initially, a single factor is added for each satellite, and a new factor is added only if the there is a cycle-slip or if a new satellite is tracked. By treating the carrier-phase bias factor in this manner (i.e., like a ``landmark'' variable in traditional pose-graph SLAM \cite{grisetti2010tutorial}), and utilizing the Bayes tree based optimizer, an efficient real-time smoothing formulation for GNSS signal processing is presented. The measurement Jacobian associated with this graph construction is represented in Fig. \ref{fig:jacobian}.B. From Fig. \ref{fig:jacobian}.B, we see a less densely populated measurement Jacobian, as desired.

\subsection{Incremental Factor Graph Inference}

The formulation presented in the previous sections provides an efficient estimation of $X^{MAP}$ when all of the information is provided $a \ priori$. However, it is generally the case that information is arriving sequentially, and it is desired to incrementally provide state estimates. The ability to provide an incremental estimator lies in the capability of the optimizer to reuse prior computations. A well studied technique of computation reuse, for state estimation, is to employ QR-factorization to update the previous matrix factorization \cite{bierman2006factorization, isam}; however, this technique only works for linearized systems.

To overcome this limitation, the Incremental Smoothing and Mapping (iSAM2) formulation was developed ~\cite{isam2}. The iSAM2 formulation allows for incremental inference over linear or non-linear objective functions through the utilization of a novel graphical model, the Bayes tree ~\cite{bayesTree}. To provide insight into this formulation, specifically for GNSS applications, a simple GNSS example will be presented. Where it will be shown how to convert the GNSS factor graph into a Bayes tree. Additionally, a discussion will be provided on how the Bayes tree graphical models allows for efficient inference.

\begin{figure}[htb!]
 \begin{center}
  \begin{adjustbox}{width=0.45\textwidth}
 \begin{tikzpicture}[
  rednode/.style={shape=rectangle, fill=red},
  bluenode/.style={shape=rectangle, fill=blue},
  greennode/.style={shape=rectangle, fill=green},
  yellownode/.style={shape=rectangle, fill=yellow}
  ]
  \node[shape=rectangle, fill=red, minimum size=0.5cm] (Z) at (2,0) {};

  \node[shape=circle,fill=grey, minimum size=1.5cm] (B) at (4,0) {$X_{1}$};

  \node[shape=rectangle,fill=blue, minimum size=0.5cm] (H) at (4,2) {};

  \node[shape=rectangle,fill=green, minimum size=0.5cm] (Q) at (6,0) {};

  \node[shape=rectangle, fill=yellow, minimum size=0.5cm] (S) at (5,-2) {};

  \node[shape=rectangle, fill=yellow, minimum size=0.5cm] (T) at (7,-2) {};

  \node[shape=circle,fill=grey, minimum size=1.5cm] (E) at (6,-4) {$B_{1}$};

  \node[shape=rectangle, fill=red, minimum size=0.5cm] (X) at (8,-4) {};

  \node[shape=circle,fill=grey, minimum size=1.5cm] (C) at (8,0) {$X_{2}$};

  \node[shape=rectangle,fill=blue, minimum size=0.5cm] (L) at (8,2) {};

  \node[shape=rectangle,fill=green, minimum size=0.5cm] (R) at (10,0) {};

  \node[shape=circle,fill=grey, minimum size=1.5cm] (D) at (12,0) {$X_{3}$};

  \node[shape=circle,fill=grey, minimum size=1.5cm] (F) at (12,-4) {$B_{2}$};

  \node[shape=rectangle, fill=blue, minimum size=0.5cm] (W) at (12,2) {};

  \node[shape=rectangle, fill=yellow, minimum size=0.5cm] (U) at (12,-2) {};

  \node[shape=rectangle, fill=red, minimum size=0.5cm] (V) at (10,-4) {};

  \path [-,line width=2pt] (Z) edge node {} (B);
  \path [-,line width=2pt] (B) edge node {} (H);

  \path [-,line width=2pt] (B) edge node {} (Q);
  \path [-,line width=2pt] (Q) edge node {} (C);

  \path [-,line width=2pt] (C) edge node {} (L);
  \path [-,line width=2pt] (C) edge node {} (R);

  \path [-,line width=2pt] (D) edge node {} (R);

  \path [-,line width=2pt] (D) edge node {} (U);
  \path [-,line width=2pt] (U) edge node {} (F);

  \path [-,line width=2pt] (B) edge node {} (S);
  \path [-,line width=2pt] (S) edge node {} (E);
  \path [-,line width=2pt] (E) edge node {} (X);

  \path [-,line width=2pt] (W) edge node {} (D);
  \path [-,line width=2pt] (C) edge node {} (T);
  \path [-,line width=2pt] (T) edge node {} (E);
  \path [-,line width=2pt] (F) edge node {} (V);

  \matrix [draw, above] at (current bounding box.south west) {
   \node [rednode,label=right:Prior Factors] {}; \\
   \node [greennode,label=right:Between Factors] {}; \\
   \node [bluenode,label=right:Pseudorange Factors] {}; \\
   \node [yellownode,label=right:Carrier Phase Factors] {}; \\
  };

 \end{tikzpicture}
\end{adjustbox}
 \end{center}
 \caption{GNSS factor graph construction.}
 \label{fig:gnssGraph}
\end{figure}
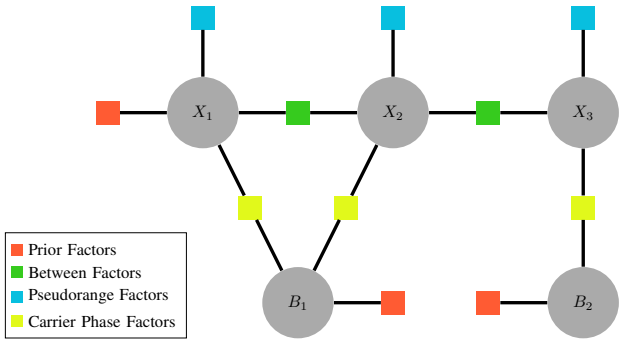

To begin our discussion, a factor graph that represents the GNSS inference problem is presented in Fig. \ref{fig:gnssGraph}. With this factor graph, it is desired to estimate the states $\{X,B\}$. In this formulation,  X represents the position, troposphere, and receiver clock bias states, as provided in Eq. \ref{eq:stateVec}. Additionally, the vertices B represents the carrier-phase bias states.

\begin{equation}
 \label{eq:stateVec}
 X = \begin{pmatrix}
 \delta P    \\
 T_{z,w}   \\
 C_b \\
 \end{pmatrix}
\end{equation}

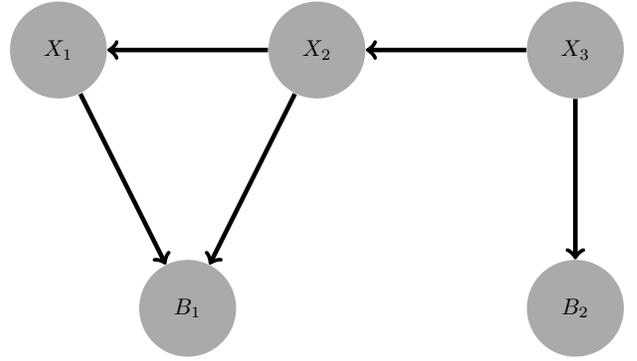
\begin{figure}[htb!]
 \begin{center}
  \begin{adjustbox}{width=0.45\textwidth}
 \begin{tikzpicture}

  \node[shape=circle,fill=grey, minimum size=1.5cm] (A) at (4,0) {$X_{1}$};
  \node[shape=circle,fill=grey, minimum size=1.5cm] (B) at (6,-4) {$B_{1}$};
  \node[shape=circle,fill=grey, minimum size=1.5cm] (C) at (8,0) {$X_{2}$};
  \node[shape=circle,fill=grey, minimum size=1.5cm] (D) at (12,0) {$X_{3}$};
  \node[shape=circle,fill=grey, minimum size=1.5cm] (E) at (12,-4) {$B_{2}$};

  \path [->,line width=2pt] (A) edge node {} (B);
  \path [->,line width=2pt] (C) edge node {} (A);
  \path [->,line width=2pt] (C) edge node {} (B);
  \path [->,line width=2pt] (D) edge node {} (C);
  \path [->,line width=2pt] (D) edge node {} (E);

 \end{tikzpicture}
\end{adjustbox}
 \end{center}
 \caption{Generating a Bayes Net from the original factor graph using the elimination ordering, \{$B_1$, $B_2$, $X_1$, $X_2$, $X_3$\}}
 \label{fig:bayesNet}
\end{figure}

Utilizing the factor graph presented in Fig. \ref{fig:gnssGraph}, we can begin the process of converting a factor graph into a Bayes tree. To do this, we must take an intermediate step and construct a Bayes net. The Bayes net can be constructed from the factor graph using a variable elimination game \cite{heggernes1996finding}. For our specific example, the Bayes net is provided in Fig. \ref{fig:bayesNet}, if the bias states are eliminated first then the positing states (i.e., the variable elimination is ${B_1,B_2,X_1,X_2,X_3}$ ). It should be noted that if the elimination ordering is varied, the resultant Bayes net will change and this can have a substantial impact on the run-time of the optimizer \cite{agarwal2012variable}.

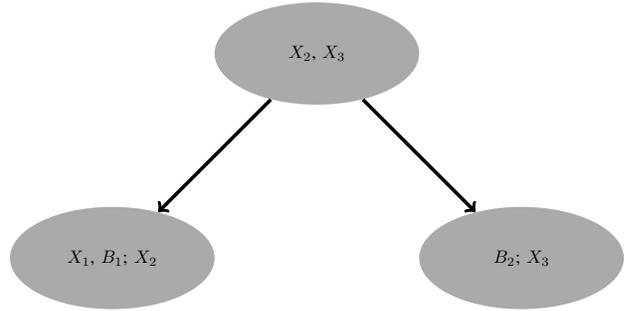
\begin{figure}[htb!]
 \begin{center}
  \begin{adjustbox}{width=0.45\textwidth}
 \begin{tikzpicture}

  \node[shape=ellipse,fill=grey, minimum height=2cm, minimum width=4cm] (A) at (4,0) {$\quad$ $X_{2}$, $X_3$ $\quad$};

  \node[shape=ellipse,fill=grey, minimum height=2cm, minimum width=4cm] (B) at (0,-4) { $X_{1}$, $B_1$; $X_2$ };

  \node[shape=ellipse,fill=grey, minimum height=2cm, minimum width=4cm] (C) at (8,-4) {$B_{2}$; $X_3$};

  \path [->,line width=2pt] (A) edge node {} (B);
  \path [->,line width=2pt] (A) edge node {} (C);

 \end{tikzpicture}
\end{adjustbox}
 \end{center}
 \caption{Generating the Bayes tree from cliques in the chordal Bayes net}
 \label{fig:bayesTree}
\end{figure}

Utilizing the previously constructed Bayes net, we can now construct the Bayes tree. The Bayes tree is constructed to take advantage of the clique structure within the Bayes net. That is, by re-writing the Bayes net we are left with a directed tree structure, where the vertices in the graph represent cliques in the original Bayes net. For our GNSS example, the constructed Bayes tree is provided in Fig. \ref{fig:bayesTree}.

The tree structure present in the Bayes tree plays a pivotal role in the ability of the data structure to provide an efficient incremental inference engine. The primary advantage of the tree structure is in the idea that only local sections (i.e., a branch in the Bayes tree structure) of the data structure needs to be re-linearized when new constraints are added to the graph. Where re-linearization can be conducted by converting a subset of the Bayes tree back into a factor graph adding the new constraint \cite{dellaert2017factor}.

\section{Experimental Setup}

\begin{figure}[htb!]
 \begin{center}
  \includegraphics[width=0.5\textwidth]{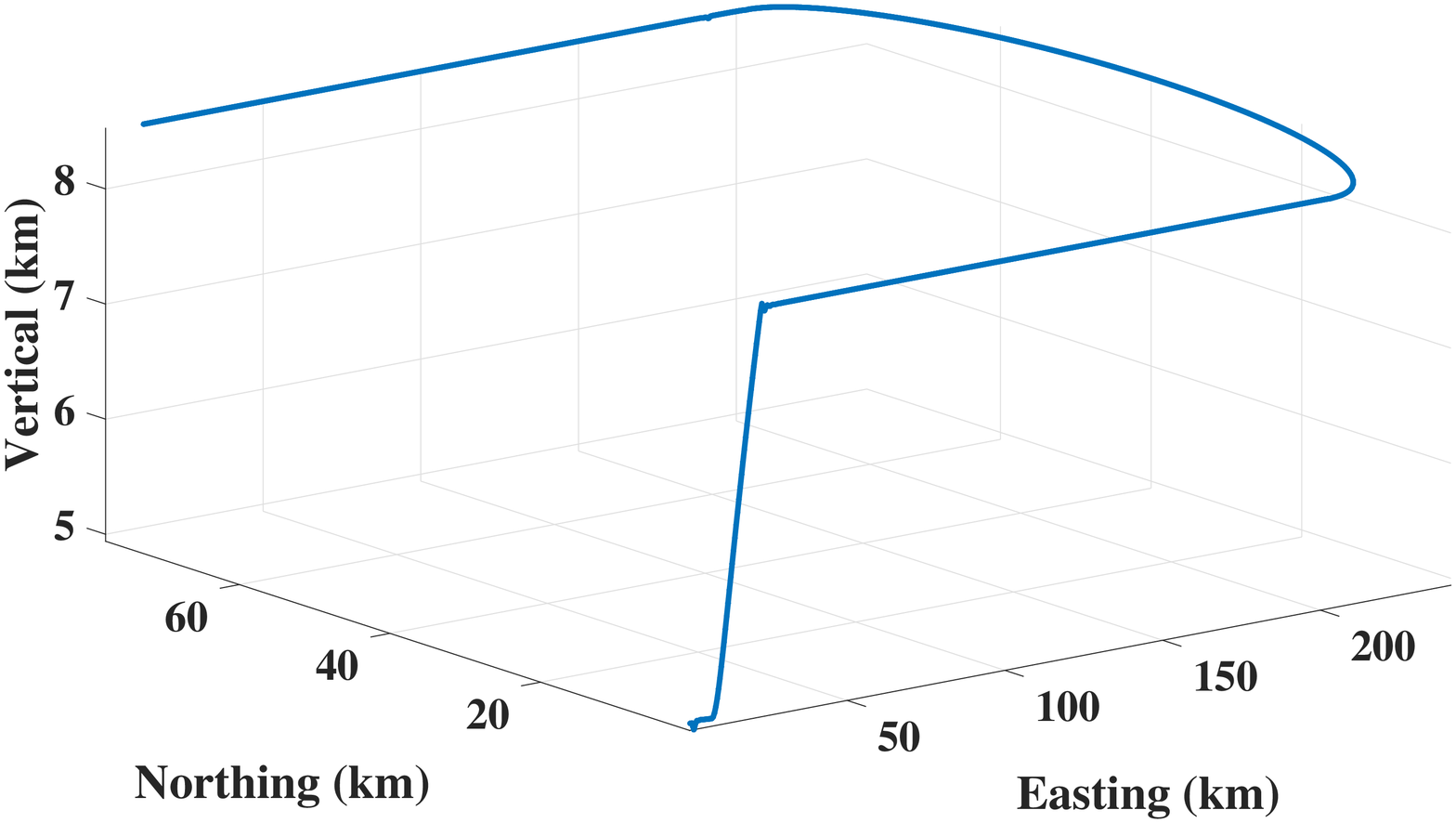}
 \end{center}
 \caption{Example flight trajectory utilized for simulated GNSS observation generation.}
 \label{fig:simProfile}
\end{figure}

To conduct an analysis of PPP convergence, a simulation environment was constructed. For synthetic observations generation, the SatNav-3.04 Toolbox ~\cite{satnavToolbox} is utilized, which provides a Matlab environment for generating dual-frequency pseudorange and carrier-phase observations for a specified trajectory. For this evaluation, four trajectories of varying dynamic were created --- an example flight trajectory is provided in Fig. \ref{fig:simProfile}. To tailor the toolbox for an evaluation of kinematic PPP airborne positioning, several minor modifications were made, as discussed in \cite{watson2016performance}.  For example, toolbox was modified to include attitude dependent satellite masking and carrier-phase breaks. That is,  when a satellite is obscured or nearly obscured due to a change in platform attitude, it is masked from view and the the potential of a carrier-phase breaks is increased. Additionally, for a PPP analysis, a methodology of constructing an orbit and clock model is required. This error model is constructed by differing JPL's International GNSS Service (IGS) submission with European Center for Orbit Determination (CODE) submission. For a more detailed discussion on the simulation environment, the reader is directed to ~\cite{watson2016performance}.

To evaluate the positioning performance of the PPP incremental graph optimizer, a Monte Carlo style experiment was implemented. Specifically, one hundred datasets were generated where several parameters, which are known to adversely affect GNSS positioning performance, were randomly initialized for each flight --- see Table \ref{table:margins} for additional information on varied parameters.

\begin{table}[htb!]
 \begin{center}
  \caption{Description of randomly initialized parameters within the Monte-Carlo evaluation.}
  \label{table:margins}
  \resizebox{\columnwidth}{!}{%
   \begin{tabular}{ | c || p{7cm} | }
    \hline

    \textbf{Varied parameters}   & \textbf{Parameter description }                            \\\hline  \hline

    \textbf{Thermal noise}       & $\sigma^{}_{\rho}=0.32m$ , $\sigma^{}_{\phi}=0.16\lambda$  \\ \hline

    \textbf{Multipath}           & $\sigma=0.4m , \tau=15sec$                                 \\ \hline

    \textbf{Tropospheric delay}  & Modified Hopfield with linear scale                        \\ \hline

    \textbf{Ionospheric delay}   & $O_{IF}$ used to mitigate error to $1^{st}$ order          \\ \hline

    \textbf{Receiver clock bias} & $\sigma= 30ns , \delta\tau^{}_{b}=100ns$                   \\ \hline

    \textbf{Phase ambiguity}     & Random initialization with attitude dependent phase breaks \\ \hline

    \textbf{Orbits}              & Orbits $\sigma = 5cm$  with linear scale                   \\ \hline
   \end{tabular}
  }
 \end{center}
\end{table}

To provide a reference positioning solution, a traditional extended Kalman filter (EKF) was utilized, where the specific EKF formulation details are provided in \cite{watson2016performance}. To provide a fair comparison, the same stochastic models were implemented for both estimators. The specific stochastic models utilized for the comparison are provided in Table \ref{table:stoch}.

\begin{table}[htb!]
 \centering
 \caption{Stochastic model parameters for estimators.}
 \resizebox{\columnwidth}{!}{%
  \begin{tabular}{|c || c | c | c  |}
   \hline
   \textbf{Parameter}              & \textbf{\textit{a priori} $\sigma$} & \textbf{Process noise}    & \textbf{ Correlation time} \\
   \hline
   \hline
   \textbf{Position}               & 1.0 $m$                             & 5 $\frac{m}{\sqrt{s}}$    & $\infty$                   \\
   \hline
   \textbf{Trop. wet zenith delay} & 0.3 $m$                             & 3e-5 $\frac{m}{\sqrt{s}}$ & $\infty$                   \\
   \hline
   \textbf{Receiver clock}         & 3e6 $m$                             & 2000 $\frac{m}{\sqrt{s}}$ & $0$                        \\
   \hline
   \textbf{Phase biases}           & $100$ $m$                           & 0 $\frac{m}{\sqrt{s}}$    & $\infty$                   \\
   \hline
  \end{tabular}
 }
 \label{table:stoch}
\end{table}

\section{Results}

\subsection{Example Simulated Flight Evaluation}

\begin{figure}[htb!]
 \begin{center}
  \includegraphics[width=0.5\textwidth]{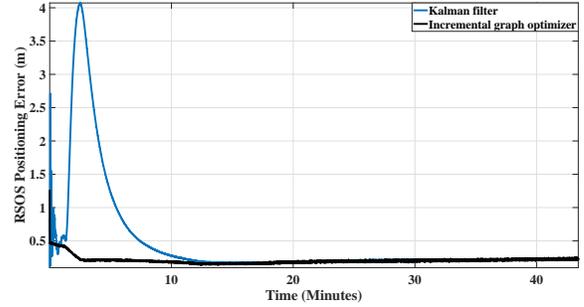}
 \end{center}
 \caption{Example RSOS positioning error profile for a typical simulated data set}
 \label{fig:singleFlightRsos}
\end{figure}

To begin an evaluation of the PPP incremental graph optimizer, a single data set, which is representative of all datasets simulated for this study, will be analyzed. As a starting point, the residual sum of squares (RSOS) positioning error is utilized to evaluated the performance of both estimators, as shown in Fig. \ref{fig:singleFlightRsos}. From Fig. \ref{fig:singleFlightRsos}, it can be readily seen that the incremental graph optimizer more quickly converges --- when compared to the EKF --- to a steady-state value. Additionally, it should be noted that the both estimators converge to approximately the same value.

The RSOS positioning error statistics for both estimator are provided in Table \ref{table:singleFlightStats}. From Table \ref{table:singleFlightStats} it should be noted that the incremental graph optimizer outperforms the EKF with respect to all metrics provided (e.g., the incremental graph optimizer provides a 25 cm error reduction with respect to the mean RSOS positioning error).

\begin{table}[htb!]
 \centering
 \caption{ Positioning statistics for a single flight }
 \resizebox{\columnwidth}{!}{%
  \begin{tabular}{|c || c | c |}
   \hline
                                   & \textbf{Incremental graph} & \textbf{Kalman filter} \\
   \hline
   \hline
   \textbf{Median} (\textbf{cm})   & 20.09                      & 21.40                  \\
   \hline
   $\mathbf{\mu}$  (\textbf{cm})   & 20.56                      & 45.68                  \\
   \hline
   $\mathbf{\sigma}$ (\textbf{cm}) & 5.13                       & 72.23                  \\
   \hline
   \textbf{Max.} (\textbf{cm})     & 126.7                      & 407.26                 \\
   \hline
  \end{tabular}
 }
 \label{table:singleFlightStats}
\end{table}

\begin{figure}[htb!]
 \begin{center}
  \includegraphics[width=0.5\textwidth]{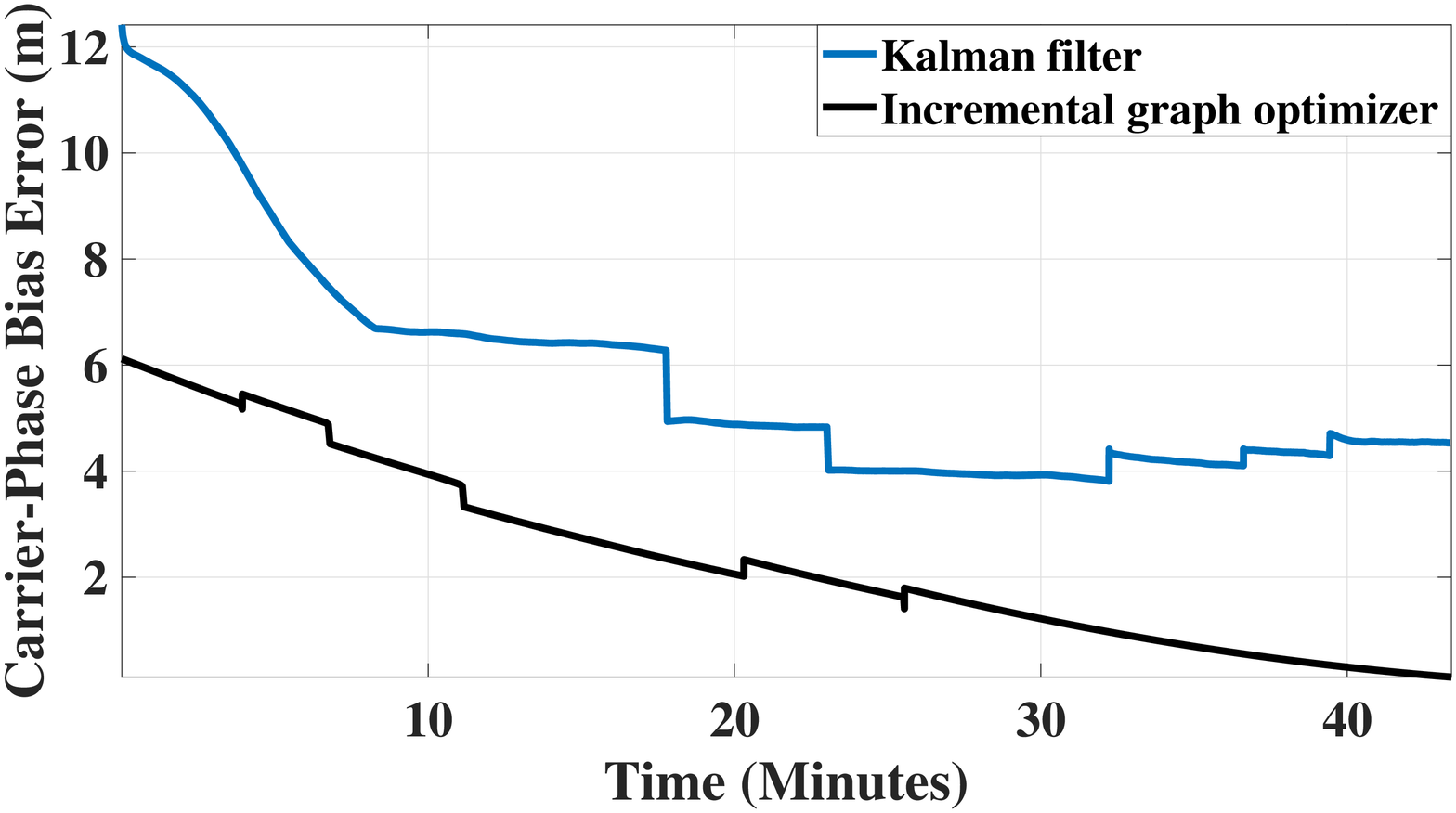}
 \end{center}
 \caption{Example phase bias convergence rate for a typical simulated data set}
 \label{fig:biasEst}
\end{figure}

To continue an analysis of this example data set, it can be seen in Fig. \ref{fig:singleFlightRsos} that the most substantial positioning error reduction  attributed to the incremental graph optimizer occurs during the first several minutes of the flight (i.e., during the PPP convergence period). To provide insight into the accelerated convergence rate of the incremental optimizer, next, an evaluation of both estimators ability to correctly resolve the phase bias states is provided in Fig. \ref{fig:biasEst}. Where it can be seen that the incremental graph optimizer provides a substantial carrier-phase bias estimation error when compared to the EKF.

\subsection{Positioning Performance Over All Simulated Flights}

Now, the evaluation shifts from a single flight to the performance of both estimators over all simulated data sets. As with the previous evaluation of a single flight, we will utilize the RSOS positioning error as the metric of comparison.

\begin{figure}[htb!]
 \begin{center}
  \includegraphics[width=0.5\textwidth]{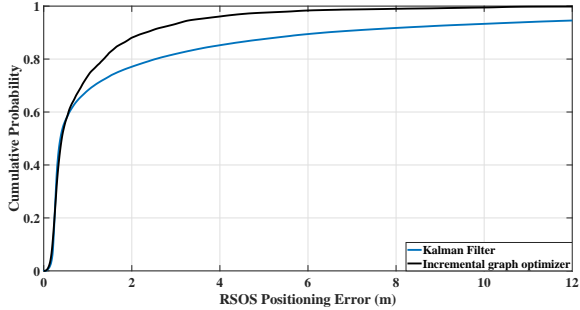}
 \end{center}
 \caption{CDF of the RSOS positioning error for all epochs over the 100 simulated datasets.}
 \label{fig:totalCdf}
\end{figure}

To begin our evaluation, the cumulative distribution function (CDF) of the RSOS positioning error for both estimators is evaluated, as provided in Fig. \ref{fig:totalCdf}. From Fig. \ref{fig:totalCdf} it can be noted that there is a considerable shift to the left for the CDF of the RSOS positioning error of the incremental graph optimizer for large error values. One possible explanation for this trend --- as indicated by our evaluation of a single flight --- is that the incremental graph optimizer is more quickly converging to a steady-state value.

\begin{table}[htb!]
 \centering
 \caption{ Positioning statistics for all epochs. }
 \resizebox{\columnwidth}{!}{%
  \begin{tabular}{|c || c | c |}
   \hline
                                   & \textbf{Incremental graph} & \textbf{Kalman filter} \\
   \hline
   \hline
   \textbf{Median} (\textbf{cm})   & 40.01                      & 39.7                   \\
   \hline
   $\mathbf{\mu}$  (\textbf{cm})   & 97.50                      & 256.27                 \\
   \hline
   $\mathbf{\sigma}$ (\textbf{cm}) & 149.46                     & 641.40                 \\
   \hline
   \textbf{Max.} (\textbf{cm})     & 2318.43                    & 10,152.28              \\
   \hline
  \end{tabular}
 }
 \label{table:totalStats}
\end{table}

\begin{figure}[htb!]
 \begin{center}
  \includegraphics[width=0.5\textwidth]{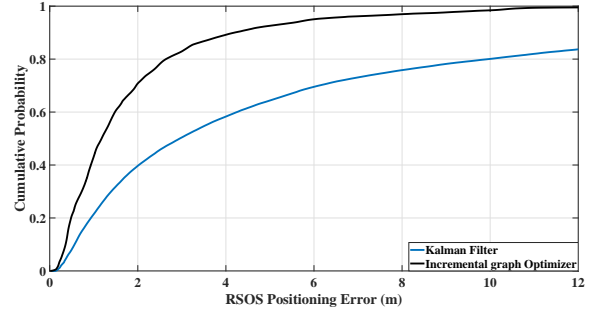}
 \end{center}
 \caption{CDF of the RSOS positioning error for epochs within the convergence period over the 100 simulated datasets.}
 \label{fig:convergeCdf}
\end{figure}

To confirm that RSOS positioning error seen in Fig. \ref{fig:totalCdf} for the incremental graph optimizer is occurring during the initial convergence period, a CDF of the RSOS positioning error for both optimizers during the first 15 minutes of each data set is provided in Fig. \ref{fig:convergeCdf}. As indicated by the right shift in Fig. \ref{fig:convergeCdf} of the EKF RSOS positioning error line, the incremental graph optimizer provides a more accurate positioning solution during the initial convergence period. The specific RSOS positioning error reduction during the initial convergence period can be seen in Table \ref{table:convergeStats}.

\begin{table}[htb!]
 \centering
 \caption{ Positioning statistics for epochs within the convergence period.}
 \resizebox{\columnwidth}{!}{%
  \begin{tabular}{|c || c | c |}
   \hline
                                   & \textbf{Incremental Graph} & \textbf{Kalman filter} \\
   \hline
   \hline
   \textbf{Median} (\textbf{cm})   & 117.76                     & 295.15                 \\
   \hline
   $\mathbf{\mu}$  (\textbf{cm})   & 188.79                     & 665.35                 \\
   \hline
   $\mathbf{\sigma}$ (\textbf{cm}) & 217.16                     & 984.01                 \\
   \hline
   \textbf{Max.} (\textbf{cm})     & 2,318.43                   & 10,152.28              \\
   \hline
  \end{tabular}
 }
 \label{table:convergeStats}
\end{table}

\subsection{Example Evaluation With Collected Data}

\begin{figure}[htb!]
 \begin{center}
  \includegraphics[width=0.5\textwidth]{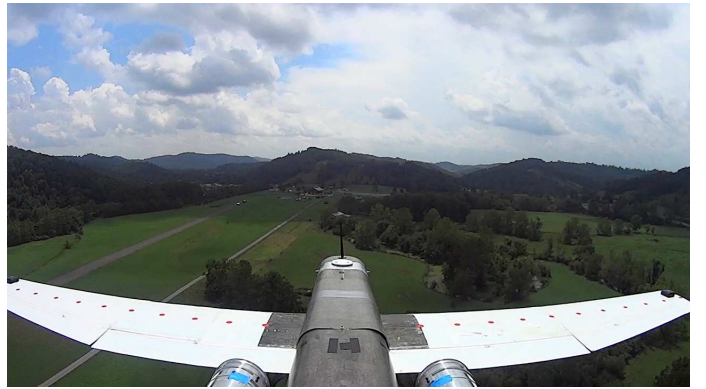}
 \end{center}
 \caption{{\it{Phastball}} research platform \cite{gu2012avionics} in flight over the West Virginia University Jackson's Mill airfield.}
 \label{fig:phastballFlight}
\end{figure}

Finally, to verify the positioning performance benefits noted in the simulation study, a similar analysis is conducted on an example collected dataset. The dataset to be evaluated was collected on-board a small, fixed-wing Unmanned Aerial Vehicle (UAV). This UAV testbed ({\it{Phastball}}) --- as depicted in Fig. \ref{fig:phastballFlight} --- was developed at West Virginia University as a research platform \cite{gu2012avionics}.

\begin{figure}[htb!]
 \begin{center}
  \includegraphics[width=0.5\textwidth]{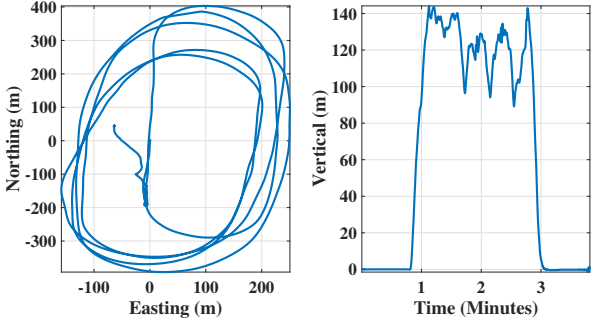}
 \end{center}
 \caption{Flight profile for collected data set.}
 \label{fig:phastballProfile}
\end{figure}

The {\it{Phastball}} is equipped with a NovAtel OEM-615 dual-frequency GNSS receiver, which provides 10 $Hz$ GNSS observables over the duration of the flight. The flight profile is depicted in Fig. \ref{fig:phastballProfile}. A second OEM-615 NovAtel GNSS receiver was placed near the runway to allow for a post-processed RTK solution, where RTKLIB \cite{takasu2011rtklib} was utilized to generated the reference solution.

\begin{figure}[htb!]
 \begin{center}
  \includegraphics[width=0.5\textwidth]{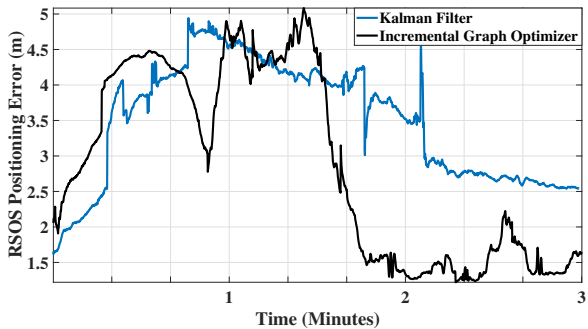}
 \end{center}
 \caption{RSOS positioning error comparison for a Kalman filter and the incremental graph optimizer when a Phastball collected dataset is utilized.}
 \label{fig:phastballError}
\end{figure}

Utilizing this dataset, the PPP incremental graph optimizer is evaluated against a Kalman filter based PPP approach, where both estimators are given the same inital conditions (i.e., both estimators are provided the same initial error covariance, the same measurement noise model, and the same process noise model). In Fig. \ref{fig:phastballError} the {\it{3D RSOS}} positioning error for both estimators is provided. The result presented in Fig. \ref{fig:phastballError} follow the trend provided by the simulation study (i.e., the incremental graph optimizer provides faster  positioning error convergence than that provided by the Kalman filter). This is further validated by looking at Table \ref{table:phastballStats}, where a substantial median {\it{3D RSOS}} positioning error reduction is granted by the PPP incremental graph optimizer when compared to the Kalman filter.

\begin{table}[htb!]
 \centering
 \caption{ Positioning statistics for collected dataset. }
 \resizebox{\columnwidth}{!}{%
  \begin{tabular}{|c || c | c |}
   \hline
                                   & \textbf{Incremental graph} & \textbf{Kalman filter} \\
   \hline
   \hline
   \textbf{Median} (\textbf{cm})   & 180.23                     & 373.56                 \\
   \hline
   $\mathbf{\mu}$  (\textbf{cm})   & 260.76                     & 349.08                 \\
   \hline
   $\mathbf{\sigma}$ (\textbf{cm}) & 127.82                     & 84.06                  \\
   \hline
   \textbf{Max.} (\textbf{cm})     & 508.32                     & 494.07                 \\
   \hline
  \end{tabular}
 }
 \label{table:phastballStats}
\end{table}

\section{Conclusion}

The desire to precisely localize a platform is of paramount importance to a myriad of fields. This desire has lead to a plethora of research into precise GNSS localization due to its ability to provide a precise and globally consistent solution. One of the most commonly utilized GNSS formulations is the precise point positioning (PPP) technique due to its autonomy from local reference stations. However, it has been noted in several studies that PPP has a relatively long initial convergence period when compared to differential techniques.

To reduce the convergence time of PPP, this paper proposes the use of recent advances in real-time smoothing made within the robotics community. Specifically, this paper makes connections between GNSS localization and incremental pose-graph optimization. The connection between the two fields lies in the ability to treat phase bias states as ``landmark'' nodes in the graph. By treating the phase bias state in this manner, and utilizing a Bayes tree based optimizer, efficient smoothing of the position states can be conducted in real-time.

To the quantify the benefit of this formulation, a Monte-Carlo style experiment was conducted within a simulation environment. Utilizing the simulated data, the incremental graph optimization was evaluated along with a traditional EKF-PPP formulation. Through this evaluation, it was found that the incremental graph optimization technique provided a substantial RSOS positioning error reduction during the initial PPP convergence period when compared to a traditional EKF formulation. This finding was also validated with an evaluation of a short duration dataset collected with a fixed-wing UAV.

Finally, to allow for external validation and collaboration, all software developed for this evaluation has been released publicly at \href{https://github.com/wvu-navLab/PPP-BayesTree}{github.com/wvu-navLab}. Included with the source code are several example datasets.

\section{Future Work}

In this evaluation, the only comparison solution was generated by a traditional EKF-PPP formulation where phase biases are estimated as floating parameters. However, there are several additional formulation that are known to provide faster convergence rates (e.g., integer ambiguity enabled PPP). With that in mind, there is a need to evaluate the incremental graph optimizer against other state of the art formulations, and to leverage these techniques within the graph.

\section*{Acknowledgment}
This work was supported in part through a subcontracts with the California Institute of Technology Jet Propulsion Laboratory and the MacAuley-Brown, Inc.  The material in this report was approved for public release on the $23^{rd}$ of February 2018, case number 88ABW-2018-0908.

\bibliography{IEEEabrv,content/phaseBias}

\end{document}